\documentclass{article}
\usepackage{spconf,amsmath,graphicx}

\usepackage{enumitem}
\setlist{nosep, leftmargin=14pt}

\usepackage{mwe} 
\usepackage{url}
\usepackage{amssymb}
\usepackage{booktabs, multirow}

\title{Reliability-Guided Vision-Language Decoding with Text-Aware Skip Refinement for Medical Image Segmentation}

\name{
Sivan Talaei$^{1}$, Fatemeh Daneshfar$^{2,*}$\thanks{$^{*}$Corresponding author: f.daneshfar@uok.ac.ir}, Abdulhady Abas Abdullah$^{3}$, Mourad Oussalah$^{4}$
}

\address{
$^{1,2}$ Department of Computer Engineering, University of Kurdistan, Sanandaj, Kurdistan, Iran \\
$^{3}$ Artificial Intelligence and Innovation Centre, University of Kurdistan, Erbil, Kurdistan, Iraq \\
$^{4}$Center for Machine Vision and Signal Analysis, University of Oulu, Oulu, Finland\\
}

\begin{document}
%
\maketitle
\begin{abstract}
Medical image segmentation remains challenging when targets exhibit limited visual distinction, irregular structure, or substantial variation across imaging modalities. Medical text can reduce this ambiguity, but indiscriminate fusion may introduce unreliable guidance. We present RELiF, a vision--language framework that integrates text through reliability-guided decoding. Its Reliability-guided Fusion Gate (RFG) adaptively controls cross-modal updates, while the Text-guided Skip Refinement Adapter (TSRA) refines high-resolution features using text guidance. RELiF is evaluated on radiographic, CT, and endoscopic segmentation tasks, where it shows competitive performance across diverse imaging characteristics and target appearances. Additional analyses examine efficiency and stability. The complementary uncertainty representation derived from the reliability response also shows an association with prediction difficulty, with higher uncertainty values occurring more frequently in pixels associated with segmentation errors. Code: \url{https://github.com/SeivanTalaie/RELiF}.

\end{abstract}

\section{Introduction}
\label{sec:introduction}
Medical image segmentation supports clinical assessment, treatment planning, and disease monitoring by identifying anatomical structures and pathological regions. Unimodal approaches, such as U-Net~\cite{ronneberger2015unet}, rely solely on visual information, which can limit their effectiveness when targets exhibit subtle appearances or substantial variation across imaging conditions. To address these challenges, multimodal approaches incorporate complementary semantic information from medical text, building on advances in vision--language learning such as CLIP~\cite{radford2021clip}. However, the benefits of textual guidance depend not only on the information it provides but also on how effectively it is integrated with visual features.

\begin{figure}[t]
    \centering
    \includegraphics[width=\columnwidth]{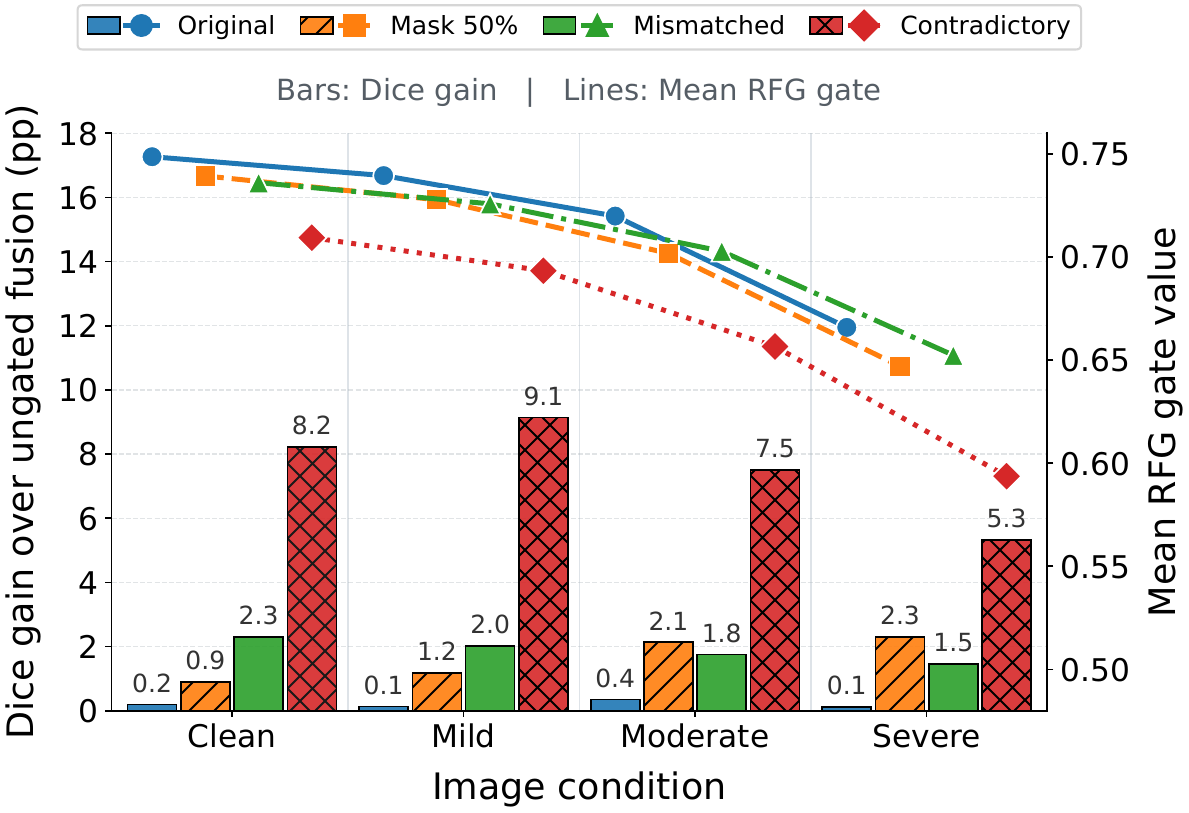}
    \caption{RFG behavior under controlled image and text degradation on QaTa-COV19. Lines show the mean gate response across the test set and three RFG stages, while bars indicate Dice gains (pp) over ungated fusion.}
    \label{fig:rfg_gate}
\end{figure} 

Recent studies have explored cross-modal reconstruction~\cite{huang2025reclmis}, uncertainty-aware learning~\cite{das2026uavls}, and bidirectional fusion~\cite{talaei2026biclip} to improve vision--language segmentation. However, how to adaptively regulate textual guidance across different visual regions during decoding remains an open question, as its relevance may vary spatially. Another limitation lies in conventional skip connections, which directly transfer high-resolution features without considering their relevance, potentially introducing domain-specific variations into the decoder~\cite{wilm2024rethinking}. These challenges call for selective cross-modal fusion and context-aware refinement of skip features.

\begin{figure*}[t]
    \centering
    \includegraphics[width=0.79\textwidth]{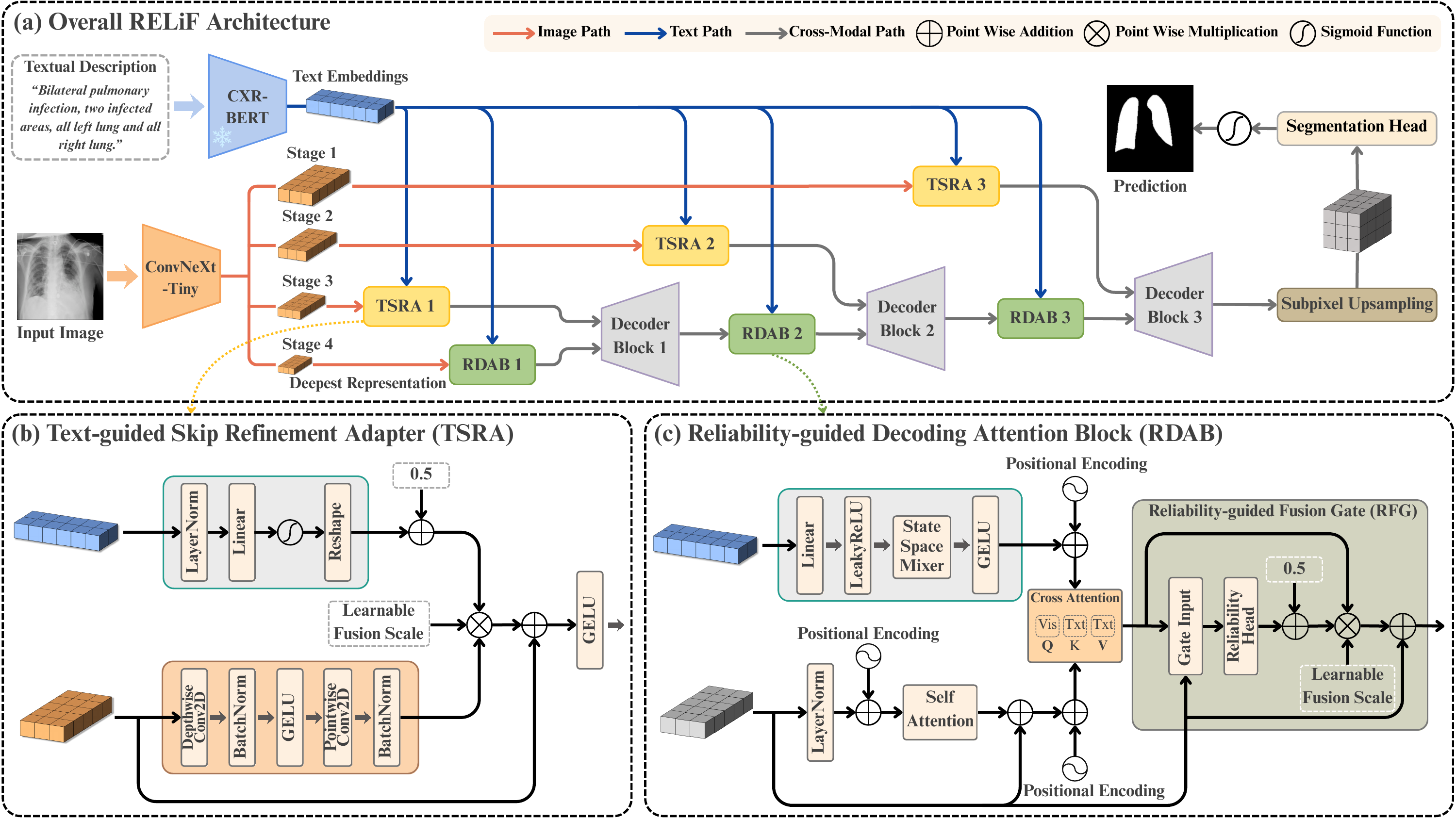}
    \caption{Overview of RELiF: (a) vision--language segmentation architecture. (b) Text-guided Skip Refinement Adapter (TSRA). (c) Reliability-guided Decoding Attention Block (RDAB) with its Reliability-guided Fusion Gate (RFG).}
    \label{fig:relif_architecture}
\end{figure*} 

To address these challenges, we propose RELiF, a vision--language framework that combines reliability-guided decoding with text-guided skip refinement. Its Reliability-guided Fusion Gate (RFG) estimates token-wise reliability from visual and text-conditioned features, together with their feature-level agreement and discrepancy, to adaptively regulate cross-modal contributions during decoding. As shown in Fig.~\ref{fig:rfg_gate}, RFG adapts its fusion response to variations in image quality and textual relevance, with corresponding Dice gains over ungated fusion. In parallel, the Text-guided Skip Refinement Adapter (TSRA) uses textual context to refine high-resolution skip features before decoder fusion. The main contributions are as follows:





\begin{itemize}
    \item We introduce RELiF, whose Reliability-guided Fusion Gate (RFG) within the Reliability-guided Decoding Attention Block (RDAB) adaptively regulates cross-modal fusion using token-wise reliability derived from visual and text-conditioned features and their feature-level similarities and differences.

    \item We develop the Text-guided Skip Refinement Adapter (TSRA) to preserve fine structural details through text-conditioned channel modulation and lightweight refinement of high-resolution skip features.

    \item We evaluate RELiF on chest X-ray and lung CT infection segmentation, as well as endoscopic polyp segmentation, demonstrating its effectiveness across diverse imaging modalities and target appearances.


    \item We conduct additional experiments and analyses to assess RELiF's effectiveness,
    reliability, efficiency, and stability.
\end{itemize}

\section{Methodology}

\subsection{Overview}

As shown in Fig.~\ref{fig:relif_architecture}, visual and textual features guide progressive decoding. RDABs regulate cross-modal fusion, while TSRAs operate on separate skip pathways, refining encoder features before their integration into decoder blocks. Finally, subpixel upsampling restores the spatial resolution, and the segmentation head generates the final pixel-wise predictions.

\subsection{Reliability-guided Fusion Gate (RFG)}
\label{subsec:rfg}

Given medical images $\mathbf{I}\in\mathbb{R}^{B\times C_I\times H\times W}$ and tokenized descriptions $\mathbf{T}$, ConvNeXt-Tiny \cite{liu2022convnext} extracts visual features $\mathbf{I}'_s$ across four encoder stages, $s\in\{1,2,3,4\}$, while BioViL CXR-BERT \cite{cxrbert} generates text embeddings $\mathbf{T}'\in\mathbb{R}^{B\times L\times D_T}$. Here, $B$ denotes batch size, $C_I$ the image channels, $H,W$ the spatial dimensions, and $L,D_T$ the text length and embedding dimension. Decoding starts from the deepest visual features $\mathbf{I}'_4$, reshaped into $\mathbf{V}\in\mathbb{R}^{B\times N_4\times C_4}$, where $N_4~=~H_4W_4$ is the number of visual tokens. Meanwhile, text embeddings pass through a learnable linear layer, LeakyReLU, SSMixer, and GELU to produce $\mathbf{T}_p\in\mathbb{R}^{B\times L_p\times C_4}$, where $L_p$ is the resulting sequence length. SSMixer models dependencies among textual tokens to provide context-aware representations at the current visual scale. Within RDAB, normalized and position-encoded visual tokens undergo multi-head self-attention. The output is normalized and added to $\mathbf{V}$ through a residual connection, yielding contextual features $\mathbf{V}_c$. Cross-attention then uses normalized, position-encoded $\mathbf{V}_c$ as queries, position-encoded $\mathbf{T}_p$ as keys, and $\mathbf{T}_p$ as values to generate text-conditioned features $\mathbf{C}$. To assess their consistency with visual information, RFG combines both representations and their feature-level relationships:
\begin{equation}
\mathbf{A}=\operatorname{Concat}\left(
\mathbf{V}_c,\mathbf{C},
|\mathbf{V}_c-\mathbf{C}|,
\mathbf{V}_c\odot\mathbf{C}
\right).
\end{equation}

A lightweight reliability head then estimates a response for each visual token:
\begin{equation}
\mathbf{R}=\sigma\left(
\operatorname{Linear}\left(
\operatorname{GELU}\left(
\operatorname{Linear}(\operatorname{LN}(\mathbf{A}))
\right)\right)\right).
\end{equation}

The complementary uncertainty representation is defined as $\mathbf{U}=1-\mathbf{R}$, with higher values indicating lower estimated reliability. The reliability response subsequently regulates cross-modal fusion through
\begin{equation}
\mathbf{G}=0.5+\mathbf{R},\qquad
\mathbf{Y}=\mathbf{V}+\alpha(\mathbf{G}\odot\mathbf{C}),
\end{equation}

where $\odot$ denotes element-wise multiplication, $\sigma$ the sigmoid function, and $\alpha$ a learnable scalar. The bounded gate $\mathbf{G}\in[0.5,1.5]$ regulates cross-modal contributions without entirely suppressing or excessively amplifying them.

\subsection{Text-guided Skip Refinement Adapter (TSRA)}
\label{subsec:tsra}

While RFG regulates cross-modal fusion, TSRA refines high-resolution skip features $\mathbf{I}'_s$ for $s\in\{1,2,3\}$ to preserve relevant spatial details. The shared text embedding $\mathbf{T}'$ is averaged across $L$ tokens to obtain $\mathbf{t}_g$, which is normalized and projected into a stage-specific channel response. A sigmoid activation and a $0.5$ shift are then applied, followed by reshaping to enable channel-wise modulation of the spatial skip features, yielding $\overline{\mathbf{g}}_s\in[0.5,1.5]^{B\times C_s\times 1\times 1}$:
\begin{equation}
\overline{\mathbf{g}}_s=\operatorname{Reshape}\left(
0.5+\sigma\left(
\operatorname{Linear}(\operatorname{LN}(\mathbf{t}_g))
\right)\right).
\end{equation}

In parallel, a depthwise $3\times3$ convolution extracts local spatial details from $\mathbf{I}'_s$, followed by batch normalization and GELU. A pointwise $1\times1$ convolution then combines information across channels, followed by another normalization to produce $\mathbf{D}_s$. The text-conditioned gate $\overline{\mathbf{g}}_s$ modulates these features through a scaled residual connection:
\begin{equation}
\widetilde{\mathbf{I}}'_s=
\operatorname{GELU}\left(
\mathbf{I}'_s+
\beta\left(
\overline{\mathbf{g}}_s\odot\mathbf{D}_s
\right)\right) \in\mathbb{R}^{B\times C_s\times H_s\times W_s}.
\end{equation}

Here, $\beta$ is a learnable residual scale, and $\odot$ denotes element-wise multiplication. Zero-initialized gate projections and small initial residual scales allow gradual refinement. Finally, the refined skip features $\widetilde{\mathbf{I}}'_s$ are fused with the upsampled decoder features without additional supervision.

\subsection{Decoder Block}
\label{subsec:decoder}

The decoder progressively restores spatial details through three stages, $s\in\{4,3,2\}$. At each stage, RDAB uses the shared text embedding $\mathbf{T}'$ to produce reliability-guided visual tokens $\mathbf{Y}_s\in\mathbb{R}^{B\times N_s\times C_s}$, where $N_s=H_sW_s$ denotes the token count and $C_s$ the channel dimension. These tokens are reshaped into spatial features $\mathbf{Z}_s\in\mathbb{R}^{B\times C_s\times H_s\times W_s}$ and upsampled through transposed convolution. The resulting features are concatenated with the corresponding TSRA-refined skip features $\widetilde{\mathbf{I}}'_{s-1}$ and convolved:
\begin{equation}
\mathbf{D}_{s-1}=\operatorname{ConvBlock}\left(
\operatorname{Concat}\left(
\operatorname{UpConv}(\mathbf{Z}_s),
\widetilde{\mathbf{I}}'_{s-1}
\right)\right)
\end{equation}

The fused features are converted back into visual tokens for the next RDAB, progressively incorporating finer spatial information. Finally, subpixel upsampling restores the input resolution, and a sigmoid-activated segmentation head generates pixel-wise foreground probabilities:
\begin{equation}
\widehat{\mathbf{M}}=\sigma\left(
\operatorname{SegHead}\left(
\operatorname{SubPixelUp}(\mathbf{D}_1)
\right)\right).
\end{equation}


\subsection{Objective Function}
\label{subsec:objective}

RELiF is trained using the Spectral-Entropic Uncertainty (SEU) objective \cite{das2026uavls}, which combines three complementary loss terms. Dice loss maximizes overlap between the predicted mask $\widehat{\mathbf{M}}$ and ground truth $\mathbf{M}$, while spectral loss encourages structural consistency by comparing their Fourier magnitude spectra. An entropy-inspired term further promotes confident foreground predictions. The overall objective is defined as
\begin{equation}
\mathcal{L}_{\mathrm{total}}
=
\mathcal{L}_{\mathrm{Dice}}
+
\lambda_F\mathcal{L}_{\mathrm{Spectral}}
+
\lambda_E\mathcal{L}_{\mathrm{Entropy}},
\end{equation}

where $\lambda_F$ and $\lambda_E$ control the contributions of the spectral and entropy terms, respectively. The entropy-inspired term is applied only to the final segmentation probabilities and is separate from the RFG-derived uncertainty representation. RFG and TSRA are learned end-to-end without direct supervision.

\section{Experiments and Results}

\subsection{Datasets}
\label{subsec:datasets}

Experiments were conducted on three public datasets: QaTa-COV19 \cite{Qata1}, MosMedData+ \cite{morozov2020mosmeddata}, and Kvasir-SEG \cite{jha2019kvasir}, covering chest X-ray, CT, and endoscopic segmentation. Following the LViT protocol \cite{li2024lvit}, we adopt the same dataset splits and use text annotations from \cite{li2024lvit} for infection segmentation and \cite{poudel2023exploring} for polyp segmentation. The dataset splits are:\\
QaTa-COV19: Training=5716; Validation=1429; Test=2113\\
MosMedData+: Training=2183; Validation=273; Test=273\\
Kvasir-SEG: Training=800; Validation=100; Test=100.

\subsection{Implementation Details}
\label{subsec:implementation_details}

The framework was implemented in PyTorch, with each dataset trained independently on an NVIDIA RTX 4090 GPU. All input images and masks were resized to $224\times224$ pixels. Training was conducted for 200 epochs, using batch sizes of 32 for training and 8 for validation and testing. We used the AdamW optimizer with a cosine annealing learning-rate scheduler. Initial learning rates were $3\times10^{-4}$ for QaTa-COV19 and Kvasir-SEG and $5\times10^{-4}$ for MosMedData+. The SEU loss weights were set to $\lambda_F=0.3$ and $\lambda_E=0.1$.

\begin{table}[t]
	\centering
	\caption{Comparison with state-of-the-art methods. best and second-best results are \textbf{bold} and \underline{underlined}, respectively.}
	\label{tab:sota_comparison}
    \vspace{5pt}
	\resizebox{\columnwidth}{!}{%
		\begin{tabular}{lcccccc}
			\toprule
			\multirow{2}{*}{Method} &
			\multicolumn{2}{c}{QaTa-COV19} &
			\multicolumn{2}{c}{MosMedData+} &
			\multicolumn{2}{c}{Kvasir-SEG} \\
            \cmidrule(lr){2-3}
            \cmidrule(lr){4-5}
            \cmidrule(lr){6-7}
			&
			Dice(\%) & mIoU(\%) &
			Dice(\%) & mIoU(\%) &
			Dice(\%) & mIoU(\%) \\
			\midrule
			
            \multicolumn{7}{l}{\textit{Monomodal models}} \\
			U-Net ('15)~\cite{ronneberger2015unet}
			& 78.91
			& 69.32
			& 64.30
			& 50.50
			& 82.33
			& 74.26 \\
				
			nnUNet ('21)~\cite{isensee2021nnunet}
			& 80.30
			& 70.62
			& 72.32
			& 60.14
			& 84.22
			& 75.41 \\
			
			UCTransNet ('22)~\cite{wang2022uctransnet}
			& 79.00
			& 69.34
			& 65.71
			& 52.55
			& 91.04
			& \underline{87.31} \\

			\midrule
			
            \multicolumn{7}{l}{\textit{Multimodal models}} \\
			
			LViT ('23)~\cite{li2024lvit}
			& 83.40
			& 74.89
			& 74.32
			& 61.03
			& 87.59
			& 75.16 \\
			
			RecLMIS ('25)~\cite{huang2025reclmis}
			& 84.93
			& 76.86
			& 77.26
			& 64.95
			& 86.58
			& 77.08 \\
			
			ProLearn ('25)~\cite{ye2025prolearn}
			& \underline{90.60}
			& \underline{82.81}
			& 77.82
			& 63.69
			& 90.40
			& 82.80 \\

            UA-VLS ('26)~\cite{das2026uavls}
			& 90.11
			& 82.00
			& 78.48
			& 64.58
			& \underline{92.41}
			& 85.89 \\

			BiCLIP ('26)~\cite{talaei2026biclip}
			& \underline{90.60}
			& \underline{82.81}
			& \textbf{81.00}
			& \textbf{68.06}
			& 89.02
			& 80.21 \\

            \midrule
			\textbf{RELiF (Ours)}
			& \textbf{91.54}
			& \textbf{84.41}
			& \underline{79.27}
			& \underline{65.66}
			& \textbf{93.28}
			& \textbf{87.41} \\
			
			\bottomrule
		\end{tabular}}
\end{table}

\begin{figure}[!t]
    \centering
    \includegraphics[width=\columnwidth]{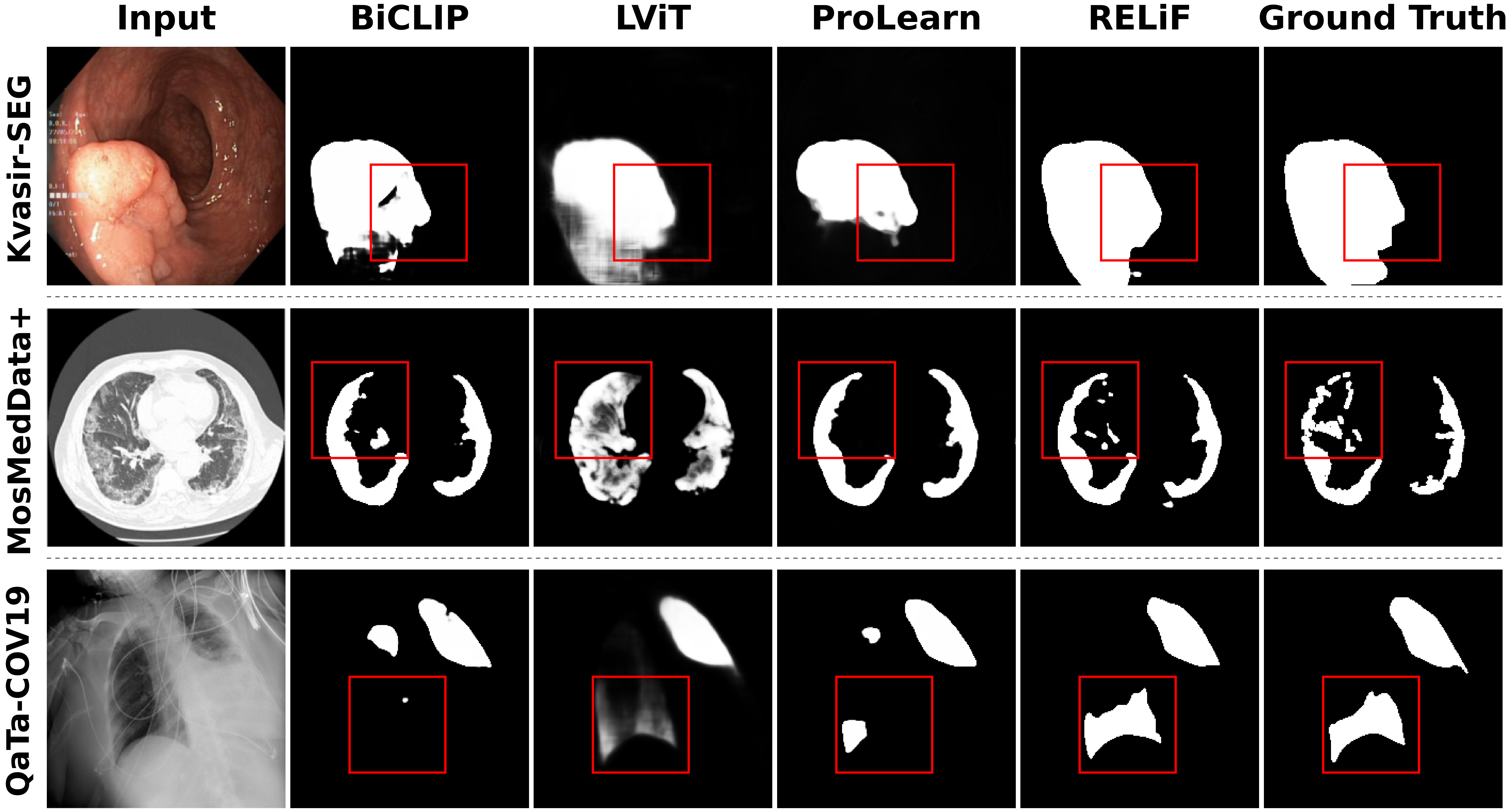}
    \caption{Qualitative segmentation results across three datasets.}
    \label{fig:qualitative_comparison}
\end{figure}

\begin{table}[!t]
	\centering
	\caption{Segmentation performance with 50\% and 10\% of the training data across three datasets.}
    \vspace{5pt}
	\label{tab:Supervision_comparison}
    \resizebox{\columnwidth}{!}{
	\begin{tabular}{lccccccc}
		\toprule
		\multirow{2}{*}{Method} &
		\multirow{2}{*}{Regime} &
		\multicolumn{2}{c}{QaTa-COV19} &
		\multicolumn{2}{c}{MosMedData+} &
		\multicolumn{2}{c}{Kvasir-SEG} \\
		\cmidrule(lr){3-4}
		\cmidrule(lr){5-6}
		\cmidrule(lr){7-8}
		& & Dice(\%) & mIoU(\%) & Dice(\%) & mIoU(\%) & Dice(\%) & mIoU(\%) \\
		\midrule
		
		
		\multirow{2}{*}{BiCLIP \cite{talaei2026biclip}}
		& 50\% & 89.83 & 81.53 & \textbf{79.72} & \textbf{66.27} & 86.55 & 76.29 \\
		& 10\% & 87.16 & 77.25 & 69.23 & 52.95 & 71.72 & 55.91 \\
		\midrule
		
		\multirow{2}{*}{LViT \cite{li2024lvit}}
		& 50\% & 78.81 & 69.55 & 58.98 & 47.28 & 80.66 & 73.30 \\
		& 10\% & 74.72 & 64.99 & 51.07 & 39.59 & 55.68 & 44.38 \\
		\midrule
		
		\multirow{2}{*}{\textbf{RELiF}}
		& 50\% & \textbf{91.10} & \textbf{83.66} & 78.06 & 64.01 & \textbf{91.95} & \textbf{85.10} \\
		& 10\% & \textbf{88.29} & \textbf{79.04} & \textbf{74.72} & \textbf{59.64} & \textbf{85.17} & \textbf{74.17} \\
		\bottomrule
	\end{tabular}}
\end{table}

\subsection{Experimental Results and Analysis}

Table~\ref{tab:sota_comparison} compares RELiF with representative monomodal and multimodal methods across three medical segmentation datasets. RELiF achieves the highest Dice and mIoU scores on QaTa-COV19 and Kvasir-SEG, while maintaining competitive performance on MosMedData+ with only 18.1 G FLOPs. Qualitative comparisons in Fig.~\ref{fig:qualitative_comparison} further illustrate the segmentation results across all three datasets. To assess robustness under limited supervision, Table~\ref{tab:Supervision_comparison} compares RELiF with BiCLIP and LViT using 50\% and 10\% training data. RELiF achieves the best Dice and mIoU on two datasets at 50\% and all three at 10\%. We further evaluated stability across five independent training runs, obtaining Dice standard deviations of 0.12, 0.14, and 0.24 percentage points on QaTa-COV19, MosMedData+, and Kvasir-SEG, respectively, indicating low run-to-run variability. The ablation study in Table~\ref{tab:component_ablation} confirms the complementary contributions of RFG and TSRA, with their combination achieving the best performance. Furthermore, removing textual guidance or introducing mismatched text during training reduces segmentation accuracy, highlighting the importance of relevant textual information. Table~\ref{tab:error_detection} further evaluates the ability of RFG-derived uncertainty to distinguish erroneous pixels from correct predictions, reporting AUROC values of 0.7724--0.8473 and AUPRC scores 3.04--5.51 times their prevalence-based random baselines. These results suggest consistent error-discrimination ability across the three datasets. Finally, we conducted a pixel-level risk--coverage analysis on Kvasir-SEG by progressively excluding high-uncertainty pixels. At 90\% coverage, retained-pixel error risk decreased by 45.2\%, with selective Dice reaching 93.87\%. Further reducing coverage to 70\% increased risk reduction to 71.7\%, although selective Dice declined, potentially due to the exclusion of correctly predicted foreground pixels.

\begin{table}[!t]
\centering
\caption{Uncertainty-based pixel error detection using RFG-derived uncertainty response. Error prevalence defines the random AUPRC baseline, while AUPRC lift measures gains.}
\vspace{5pt}
\label{tab:error_detection}
\resizebox{\linewidth}{!}{
\begin{tabular}{lccccc}
\toprule
Dataset &
Error prevalence(\%) &
AUROC $\uparrow$ &
AUPRC $\uparrow$ &
Random AUPRC &
AUPRC lift $\uparrow$ \\
\midrule
QaTa-COV19  & 2.03 & 0.7724 & 0.0824 & 0.0203 & 4.06$\times$ \\
MosMedData+ & 0.65 & 0.8389 & 0.0199 & 0.0065 & 3.04$\times$ \\
Kvasir-SEG  & 1.88 & 0.8473 & 0.1035 & 0.0188 & 5.51$\times$ \\
\bottomrule
\end{tabular}}
\end{table}

\begin{table}[!t]
	\centering
	\caption{Ablation study of RELiF evaluates individual components and textual guidance strategies using Dice and mIoU.}
    \vspace{5pt}
	\label{tab:component_ablation}
	\resizebox{\linewidth}{!}{
	\begin{tabular}{lcccc}
		\toprule
		\multirow{2}{*}{Method} &
		\multicolumn{2}{c}{MosMedData+} &
		\multicolumn{2}{c}{Kvasir-SEG} \\
		
		\cmidrule(lr){2-3}
		\cmidrule(lr){4-5}
		
		&
		Dice(\%) &
		mIoU(\%) &
		Dice(\%) &
		mIoU(\%) \\
		
		\midrule
		
		\textit{Component Ablation} & & & & \\
		
		Base
		& 78.48
		& 64.58
		& 92.37
		& 85.82 \\
		
		Base + RFG
		& 78.92
		& 65.18
		& 92.89
		& 86.72 \\
		
		Base + TSRA
		& 79.06
		& 65.37
		& 92.91
		& 86.76 \\
		
		Full RELiF (RFG + TSRA)
		& \textbf{79.27}
		& \textbf{65.66}
		& \textbf{93.28}
		& \textbf{87.41} \\
		
		\midrule
		
		\textit{Textual Guidance} & & & & \\
		
		
		Training w/o Text Guidance
		& 77.78
		& 63.64
		& 88.90
		& 80.02 \\
		
		Training w/ Mismatched Text
		& 78.73
		& 64.92
		& 91.01
		& 83.50 \\
		
		
		\bottomrule
	\end{tabular}}
\end{table}

\section{Conclusion}

This paper introduced RELiF, combining reliability-guided fusion (RFG) and text-conditioned skip refinement (TSRA) for medical segmentation. The results suggest benefits from complementary mechanisms under limited supervision. Sensitivity to mismatched text indicates that gains depend on relevant guidance rather than text alone. RFG-derived uncertainty helps identify unreliable regions, although excessive rejection may remove correctly segmented foreground. These findings motivate further reliability-aware segmentation research.

\section{Compliance with Ethical Standards}
The study retrospectively used open-access human data, with no ethical approval required per the dataset license.

\section{Acknowledgments}
This study received no funding, and the authors have no relevant conflicts of interest to disclose.

\bibliographystyle{IEEEbib}
\bibliography{strings,refs}

@inproceedings{ronneberger2015unet,
  author    = {Olaf Ronneberger and Philipp Fischer and Thomas Brox},
  title     = {{U-Net}: Convolutional Networks for Biomedical Image Segmentation},
  booktitle = {Medical Image Computing and Computer-Assisted Intervention -- MICCAI 2015},
  series    = {Lecture Notes in Computer Science},
  volume    = {9351},
  pages     = {234--241},
  publisher = {Springer},
  year      = {2015},
  doi       = {10.1007/978-3-319-24574-4_28}
}

@inproceedings{radford2021clip,
  author    = {Alec Radford and Jong Wook Kim and Chris Hallacy and
               Aditya Ramesh and Gabriel Goh and Sandhini Agarwal and
               Girish Sastry and Amanda Askell and Pamela Mishkin and
               Jack Clark and Gretchen Krueger and Ilya Sutskever},
  title     = {Learning Transferable Visual Models From Natural Language
               Supervision},
  booktitle = {Proceedings of the 38th International Conference on
               Machine Learning},
  series    = {Proceedings of Machine Learning Research},
  volume    = {139},
  pages     = {8748--8763},
  publisher = {PMLR},
  year      = {2021}
}

@article{huang2025reclmis,
  author  = {Huang, Xiaoshuang and Li, Hongxiang and Cao, Meng
             and Chen, Long and You, Chenyu and An, Dong},
  title   = {Cross-Modal Conditioned Reconstruction for Language-Guided
             Medical Image Segmentation},
  journal = {IEEE Transactions on Medical Imaging},
  volume  = {44},
  number  = {4},
  pages   = {1821--1835},
  year    = {2025},
  month   = apr,
  doi     = {10.1109/TMI.2024.3523333},
  url     = {https://doi.org/10.1109/TMI.2024.3523333}
}

@INPROCEEDINGS{das2026uavls,
  author={Das, Aryan and Rachamalla, Tanishq and Biswas, Koushik and Roy, Swalpa Kumar and Kumar Verma, Vinay},
  booktitle={2026 IEEE/CVF Winter Conference on Applications of Computer Vision (WACV)}, 
  title={Uncertainty-Aware Vision-Language Segmentation for Medical Imaging}, 
  year={2026},
  volume={},
  number={},
  pages={8490-8499},
  doi={10.1109/WACV61042.2026.00819}}

@article{talaei2026biclip,
  author        = {Saivan Talaei and Fatemeh Daneshfar and
                   Abdulhady Abas Abdullah and Mustaqeem Khan},
  title         = {{BiCLIP}: Bidirectional and Consistent Language-Image
                   Processing for Robust Medical Image Segmentation},
  journal       = {arXiv preprint arXiv:2603.00156},
  year          = {2026},
  eprint        = {2603.00156},
  archivePrefix = {arXiv},
  primaryClass  = {cs.CV},
  doi           = {10.48550/arXiv.2603.00156}
}

@article{wilm2024rethinking,
	author  = {Frauke Wilm and Jonas Ammeling and Mathias {\"O}ttl
		and Rutger H. J. Fick and Marc Aubreville
		and Katharina Breininger},
	title   = {Rethinking U-net Skip Connections for Biomedical Image
		Segmentation},
	journal = {arXiv preprint arXiv:2402.08276},
	year    = {2024},
	eprint  = {2402.08276},
	archivePrefix = {arXiv},
	primaryClass  = {eess.IV},
	doi     = {10.48550/arXiv.2402.08276}
}

@inproceedings{cxrbert,
  author    = {Boecking, Benedikt and Usuyama, Naoto and Bannur, Shruthi
               and Castro, Daniel C. and Schwaighofer, Anton
               and Hyland, Stephanie and Wetscherek, Maria
               and Naumann, Tristan and Nori, Aditya
               and Alvarez-Valle, Javier and Poon, Hoifung
               and Oktay, Ozan},
  title     = {Making the Most of Text Semantics to Improve Biomedical
               Vision--Language Processing},
  booktitle = {Computer Vision -- {ECCV} 2022},
  series    = {Lecture Notes in Computer Science},
  volume    = {13696},
  pages     = {1--21},
  publisher = {Springer Nature Switzerland},
  address   = {Cham},
  year      = {2022},
  month     = oct,
  doi       = {10.1007/978-3-031-20059-5_1}
}

@inproceedings{liu2022convnext,
  title     = {A ConvNet for the 2020s},
  author    = {Liu, Zhuang and Mao, Hanzi and Wu, Chao-Yuan and
               Feichtenhofer, Christoph and Darrell, Trevor and Xie, Saining},
  booktitle = {Proceedings of the IEEE/CVF Conference on Computer Vision
               and Pattern Recognition},
  pages     = {11976--11986},
  year      = {2022}
}

@INPROCEEDINGS{Qata1,
  author={Degerli, Aysen and Kiranyaz, Serkan and Chowdhury, Muhammad E. H. and Gabbouj, Moncef},
  booktitle={2022 IEEE International Conference on Image Processing (ICIP)}, 
  title={Osegnet: Operational Segmentation Network for Covid-19 Detection Using Chest X-Ray Images}, 
  year={2022},
  pages={2306-2310},
  doi={10.1109/ICIP46576.2022.9897412}}

@article{morozov2020mosmeddata,
  title={Mosmeddata: Chest ct scans with covid-19 related findings dataset},
  author={Morozov, Sergey P and Andreychenko, Anna E and Pavlov, Nikolay A and Vladzymyrskyy, AV and Ledikhova, Natalya V and Gombolevskiy, Victor A and Blokhin, Ivan A and Gelezhe, Pavel B and Gonchar, AV and Chernina, V Yu},
  journal={arXiv preprint arXiv:2005.06465},
  year={2020}
}

@inproceedings{jha2019kvasir,
  title={Kvasir-seg: A segmented polyp dataset},
  author={Jha, Debesh and Smedsrud, Pia H and Riegler, Michael A and Halvorsen, P{\aa}l and De Lange, Thomas and Johansen, Dag and Johansen, H{\aa}vard D},
  booktitle={International conference on multimedia modeling},
  pages={451--462},
  year={2019},
  organization={Springer}
}

@article{poudel2023exploring,
  title={Exploring transfer learning in medical image segmentation using vision-language models},
  author={Poudel, Kanchan and Dhakal, Manish and Bhandari, Prasiddha and Adhikari, Rabin and Thapaliya, Safal and Khanal, Bishesh},
  journal={arXiv preprint arXiv:2308.07706},
  year={2023}
}

@ARTICLE{li2024lvit,
  author={Li, Zihan and Li, Yunxiang and Li, Qingde and Wang, Puyang and Guo, Dazhou and Lu, Le and Jin, Dakai and Zhang, You and Hong, Qingqi},
  journal={IEEE Transactions on Medical Imaging}, 
  title={LViT: Language Meets Vision Transformer in Medical Image Segmentation}, 
  year={2024},
  volume={43},
  number={1},
  pages={96-107},
  doi={10.1109/TMI.2023.3291719}}

@article{isensee2021nnunet,
  author  = {Fabian Isensee and Paul F. Jaeger and Simon A. A. Kohl and
             Jens Petersen and Klaus H. Maier-Hein},
  title   = {{nnU-Net}: A Self-Configuring Method for Deep Learning-Based
             Biomedical Image Segmentation},
  journal = {Nature Methods},
  volume  = {18},
  number  = {2},
  pages   = {203--211},
  year    = {2021},
  doi     = {10.1038/s41592-020-01008-z}
}

@inproceedings{ye2025prolearn,
  author    = {Shuchang Ye and Usman Naseem and Mingyuan Meng and Jinman Kim},
  title     = {Alleviating Textual Reliance in Medical Language-Guided
               Segmentation via Prototype-Driven Semantic Approximation},
  booktitle = {Proceedings of the IEEE/CVF International Conference on
               Computer Vision},
  pages     = {22316--22326},
  year      = {2025}
}

@article{wang2022uctransnet,
  author  = {Wang, Haonan and Cao, Peng and Wang, Jiaqi and Zaiane, Osmar R.},
  title   = {{UCTransNet}: Rethinking the Skip Connections in {U-Net} from a Channel-Wise Perspective with Transformer},
  journal = {Proceedings of the AAAI Conference on Artificial Intelligence},
  volume  = {36},
  number  = {3},
  pages   = {2441--2449},
  year    = {2022},
  month   = jun,
  doi     = {10.1609/aaai.v36i3.20144}
}

\end{document}